\documentclass[journal]{IEEEtran}
\usepackage[pdftex]{graphicx}
\usepackage{ragged2e}                       
\usepackage{booktabs, makecell, multirow, tabularx}   
\usepackage{multirow}
\usepackage{threeparttable}
\usepackage[inline]{enumitem}

\usepackage{float}
\usepackage{caption}
\usepackage{array}
\usepackage{algorithmic}
\usepackage{subfigure}
\usepackage{xspace}
\usepackage{array}
\usepackage[export]{adjustbox}
\newcolumntype{P}[1]{>{\centering\arraybackslash}p{#1}}
\usepackage{cite}
\usepackage[font=small,labelsep=period]{caption}
\usepackage{lipsum}
\usepackage{bbding}
\usepackage{dingbat}
\usepackage{rotating}
\usepackage{dblfloatfix}
\usepackage{url}
\usepackage[table,xcdraw]{xcolor}
\usepackage[font=small,labelsep=period]{caption}
\usepackage{lipsum}
\usepackage{float}
\usepackage{comment}
\usepackage{soul}
\usepackage{url}
\usepackage{hyperref}
\usepackage{lscape}
\usepackage{academicons}
\usepackage{xcolor}
\usepackage[T1]{fontenc}

\usepackage{amsfonts}
\usepackage{fancyhdr}
\fancypagestyle{firstpage}{%
  \chead{This work has been submitted to the IEEE for possible publication. Copyright may be transferred without notice, after which this version may no longer be accessible. 
}
 
}

\title{Image Transformation for IoT Time-Series Data: A Review\\
}

\author{Duygu Altunkaya$^{1}$ , Feyza Yıldırım Okay$^{2}$ and Suat Özdemir $^{3}$
\thanks{}
\thanks{$^{1}$D. Altunkaya, Department of Computer Engineering, Konya Food and Agriculture University, Konya, e-mail: duygu.altunkaya@gidatarim.edu.tr}%
\thanks{$^{2}$F. Y. Okay, Department of Computer Engineering, Gazi University, Ankara, e-mail: feyzaokay@gazi.edu.tr}%
\thanks{$^{3}$S. Özdemir, Department of Computer Engineering, Hacettepe University, Ankara, e-mail: ozdemir@cs.hacettepe.edu.tr}%
}

\begin{document}

\maketitle
\thispagestyle{firstpage}
\begin{abstract}

In the era of the Internet of Things (IoT), where smartphones, built-in systems, wireless sensors, and nearly every smart device connect through local networks or the internet, billions of smart things communicate with each other and generate vast amounts of time-series data. As IoT time-series data is high-dimensional and high-frequency, time-series classification or regression has been a challenging issue in IoT. Recently, deep learning algorithms have demonstrated superior performance results in time-series data classification in many smart and intelligent IoT applications. However, it is hard to explore the hidden dynamic patterns and trends in time-series. Recent studies show that transforming IoT data into images improves the performance of the learning model. In this paper, we present a review of these studies which use image transformation/encoding techniques in IoT domain. We examine the studies according to their encoding techniques, data types, and application areas. Lastly, we emphasize the challenges and future dimensions of image transformation.

\end{abstract}


\section{INTRODUCTION}

The Internet of Things (IoT) refers to a network of intelligent physical devices embedded with sensors, software, and cutting-edge technologies that empower them to establish connections and share data with other devices and systems via the Internet \cite{Cc2020}. In other words, smartphones, wireless sensors, built-in systems, and nearly every device are connected and communicated via a local network or the internet. Some of the IoT applications contain smart homes \cite{BORISSOVA2022}, smart cities \cite{Aditya2023}, smart agriculture \cite{R2023}, smart health \cite{Juyal2021}, smart retail \cite{Hassan2020}, etc.

The proliferation of IoT and the growing number of IoT devices have led to the generation of immense volumes of time-series data. Consequently, time-series analysis has been performed extensively across a diverse spectrum of IoT domains \cite{Malki2022,Herrera2022}.
Traditional time-series analysis methods have accomplished convenient performance with hand-crafted characteristics and satisfactory expert knowledge. On the other hand, these methods may not always be suitable for examining IoT time-series data due to unique features that distinguish it from non-IoT time-series data \cite{Liu2023}. Analyzing time-series data for IoT devices presents challenges due to its complex nature, unlike non-IoT time-series data analysis. IoT time-series data can be quite complex, with spatial and temporal correlations that are often difficult to manage. In addition, many IoT applications require real-time or near-real-time data processing in order to make timely decisions, which can be technically challenging and require specialized infrastructure.

To address these challenges, image transformation/encoding techniques have been proposed as a promising technology by transforming time-series data into visual representations, enabling easier analysis and interpretation. In addition, transforming the data into an image and applying image compression techniques like jpeg or png can reduce the data size while preserving essential information. Compressed image representations of time-series data can be stored or transmitted more efficiently. In recent years, researchers have focused on time-series data transformation into an image format because of the great success achieved in the IoT domain, particularly in applications such as anomaly detection, fault diagnosis, and activity recognition.

In this study, we conduct a comprehensive survey of image transformation techniques from several perspectives.  Initially, we scrutinize existing studies based on their transformation techniques and subsequently categorize them according to data types (univariate or multivariate) and application domains. To the best of our knowledge, no prior survey paper has investigated the utilization of image transformation techniques in the realm of IoT. To bridge this gap, our paper presents an in-depth analysis of the current research landscape within the IoT domain.

\subsection{Motivation}

The fundamental idea of improving a model is to change it to another model which has higher accuracy. Many researchers apply combining models such as hybrid models or pre-trained models \cite{Shahin2022,Lin2022,Hong2017,Xu2020}.
However, it is worth considering whether model accuracy can be improved without altering the model itself. Some studies suggest that the transformation of time-series data may be a more effective approach for improving model accuracy than changing the model itself \cite{Yang2024,LinadWang2023,VelascoGallego2022}.

These aspects form the basis of our study's motivation. There are several advantages of representing IoT data as images:
\begin {enumerate*}[label=\itshape\roman*\upshape)]

\item It becomes easier to visualize and analyze complex patterns or trends.
\item It provides visual representations of temporal data, allowing for intuitive interpretation and pattern recognition.
\item Transforming highly dimensional IoT time-series data can be an effective way to reduce dimensionality while maintaining temporal dependencies, which can lead to more efficient analysis and better insights.
\item Deep learning techniques can be effectively employed to analyze IoT time-series data in image-based analysis for IoT applications such as pattern classification or healthcare monitoring.
\end {enumerate*} 


\subsection{Contribution}

Image transformation stands as a significant innovation with the potential to enhance outcomes not only in the realm of IoT but also across various other domains. To the best of our knowledge, there is no existing study that reviews image transformation techniques in the realm of IoT. The contributions of this paper are summarized as follows:

\begin{itemize}
    \item This study introduces the first survey paper that summarizes time-series transformation techniques in IoT. 
    \item We provide a comprehensive comparison of recent studies according to their encoding techniques, data types, and application areas in the IoT domain.
    \item We present challenges and future directions of transformation time-series into images in the context of IoT.

\end{itemize}

\subsection {Organization}
The rest of this paper is organized as follows: Section II gives in-depth information on time-series analysis in IoT and on image transformation. Section III presents image transformation techniques. A comprehensive literature review that uses time-series data in IoT applications is presented in Section IV. Section V outlines the challenges and future research directions. Lastly, Section VI concludes the paper by emphasizing key important things.

\section{Preliminaries}

\subsection{Time-Series Analysis in IoT}

A time-series is a sequence of data points collected at regular intervals over time, $X=\{(t_1,x_1), (t_2,x_2),..., (t_n,x_n)\}$,  $x_i\in\mathbb{R}^m$, where n is the number of time-series data points and m is the vector dimension. The time-series can be univariate or multivariate \cite{Chiarot2023}.

\begin{itemize}
    \item Univariate Time-Series (UTS): If $m$ equals 1, $X$ is univariate. That means UTS includes a single variable observed over time.
    \item Multivariate Time-Series (MTS): If $m$ is greater than 1, $X$ is multivariate. In other words, MTS has multiple variables observed over time.
\end{itemize}

For instance, a time-series containing the daily average temperature of a city is represented as UTS, while a time-series containing daily weather conditions (including temperature, moisture, and precipitation) for a city is represented as MTS. Although many real-world IoT systems have a large number of heterogeneous IoT sensors, there is more emphasis on UTS than MTS for some reasons. First, it is difficult to obtain the relationships between the variables in MTS correctly. Then, the fact that these variables have a high dimensionality poses a challenge when it comes to analyzing MTS data. \cite{Wan2022}. So, UTS is simpler and easier to implement than MTS. On the other hand, MTS is more complex and requires more data than UTS. However, MTS can be more accurate because it deals with relationships between different variables.

\begin{figure}
    \centering
    \includegraphics[width=0.5\textwidth]{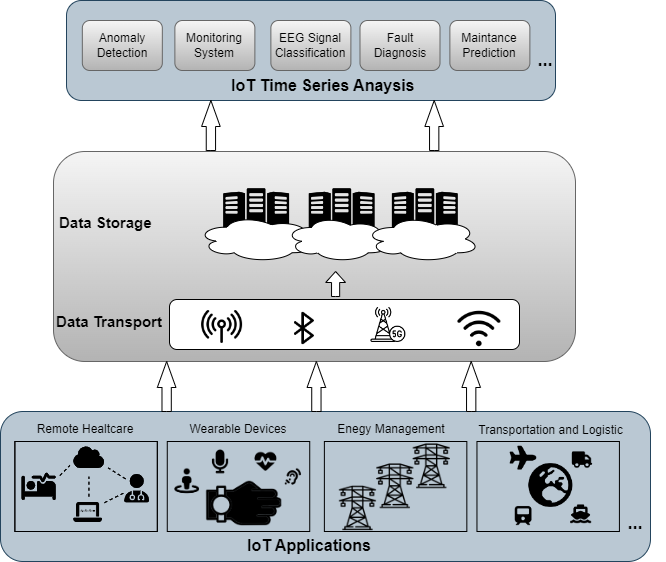}
    \captionsetup{belowskip=-8pt}
    \caption{The general structure of the time-series analysis in IoT}
    \label{fig:IoTFramework}
\end{figure}

IoT time-series data is generated from different fields, including remote healthcare, wearable devices, energy management, smart buildings, transportation, etc., as shown in Figure \ref{fig:IoTFramework}. These time-series data are widely used in various IoT problems such as anomaly detection \cite{YuandYang2022}, monitoring systems \cite{Neyja2017}, signal classification \cite{Azar2022}, fault diagnosis \cite{Chen2021}, maintenance prediction \cite{Niyonambaza2020}, etc. IoT time-series data has several unique characteristics that distinguish it from other types of data and impact the analysis and interpretation of the data \cite{Stankovic2014,Cook2019}. First, with the advancements of 5G and beyond communication technologies, time-series data from IoT devices can be massive and high-dimensional, allowing for the simultaneous monitoring of billions of devices \cite{Chettri2020}. Secondly, IoT time-series include both temporal correlations and complex spatial correlations. Thirdly, IoT time-series data can be prone to noise and missing values, which occur due to sensor failures, communication issues, data transmission problems, or errors in the measurements \cite{Yen2019}. Lastly, IoT time-series data is often generated in real-time or near real-time.

Conventional time-series analysis techniques are not directly applicable due to the features of IoT time-series data mentioned above. Understanding and leveraging these characteristics of IoT time-series data is essential for effective analysis, modeling, and decision-making in IoT applications. For instance, high dimensionality is required scalability as an important challenge for IoT time-series analysis \cite{Liu2023a}. Also, since the data is continuously produced, real-time or streaming data processing methods are required to process data flow, perform instant analysis, and timely decision-making. Furthermore, noise and missing values can diminish data quality, necessitating the use of data cleaning and preprocessing techniques to ensure data integrity. To overcome these challenges, different works have been proposed by the researchers. This paper focuses not only on the method as in other studies but also on the change in the type of time-series data and the change in methods.

\subsection{Image Transformation}

\begin{figure*}
    \centering
    \includegraphics[width=0.8\textwidth]{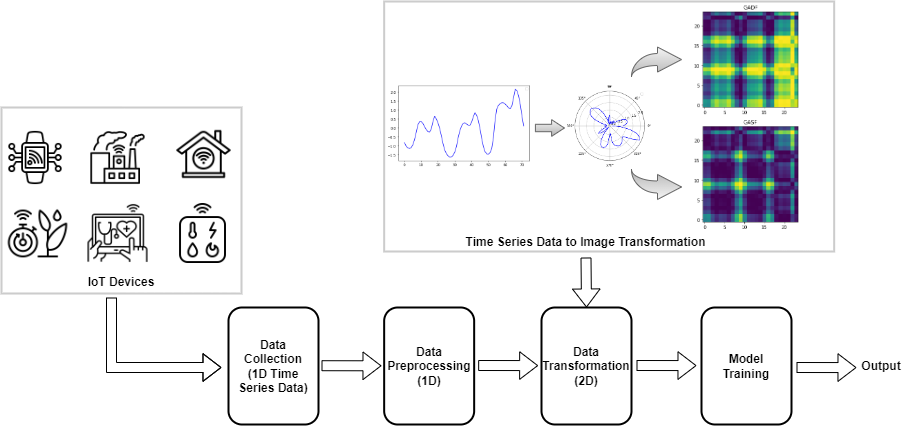}
    \caption{The overall framework of image transformation of IoT time-series data}
    \label{fig:workflow}
\end{figure*}

Time-series image transformation converts time-series data into visual representations, such as images. It is a crucial process within the IoT context. This technique reduces IoT data dimensionality by compressing extensive data into a compact visual format, making it more successful at extracting key features and patterns from IoT time-series data. Additionally, it integrates seamlessly with deep learning algorithms like Convolutional Neural Network (CNN). These transformations enhance the analysis, interpretation, and utilization of time-series data in IoT applications. The transformation process of IoT time-series data into an image is illustrated in Figure \ref{fig:workflow}.


There are varying image transformation techniques described in the literature. While these methods are directly applied to UTS, typically they are not employed directly on MTS. To address this issue, some fusion methods are discussed in the literature. Image or feature fusion is a process that is proposed to merge the necessary information from images or features \cite{Kaur2021,Sudha2017}. When converting MTS data into two-dimensional (2D) images, fusion methods can be used to combine information from different variables or data sources to create a single image representation. One of the popular fusion techniques in the literature is channel-based fusion, in which an RGB or multi-spectral channel image can be created by assigning each variable to a different color channel (e.g., red, green, blue) \cite{ChenandWang2022}. Also, some studies use tensor image fusion. MTS data is considered as a tensor and is analyzed by tensor decomposition techniques (e.g., Canonical Polyadic Decomposition) to extract patterns and interactions from the tensor data \cite{Zhou2021}. Lastly, feature level (early fusion) and decision level (late fusion)  can be utilized to transform MTS \cite{Atrey2010}. Different variables are merged at the input stage and processed together with any methods at the feature level \cite{Jiang2022}, \cite{Ehatisham-Ul-Haq2019}. On the other hand, each variable is converted into images separately, and then these images are combined at a later stage at the decision level \cite{Atrey2010}, \cite{Han2022}, \cite{Wei2020}. Also, many researchers have used hybrid fusion by performing fusion in both decision and feature levels \cite{Hang_2020}.

\section{Time-series to Image Transformation Techniques}

 There are several methods to transform one-dimensional (1D) time-series into 2D images. Some of the popular techniques in literature are discussed below. Also, Table ~\ref{table:ImageEncodingTechniques} shows the studies that used these methods.

\begin{table}
\centering

\large
\caption{The Studies of Image Transformation Techniques for IoT}
\label{table:ImageEncodingTechniques}
\resizebox{\linewidth}{!}{%
\begin{tabular}{|l|c|c|c|c|c|c|c|c|} 

\cline{3-9}
\multicolumn{1}{l}{} &\multicolumn{1}{l|}{} 
& \multicolumn{7}{c|}{\textbf{Time-Series to Image Transformation Techniques}} \\ 
\hline

\textbf{Reference} & \textbf{Year} & \multicolumn{1}{l|}{\textbf{GAF}} & \multicolumn{1}{l|}{\textbf{MTF}} & \multicolumn{1}{l|}{\textbf{RP}} & \multicolumn{1}{l|}{\textbf{STFT}} & \multicolumn{1}{l|}{\textbf{CWT}} & \multicolumn{1}{l|}{\textbf{HHT}}  & \multicolumn{1}{l|}{\textbf{Others}}  \\ 
\hline

Baldini et al. \cite{Baldini2019} & 2018 &  &  & \checkmark &  &  &  &  \\ 
\hline
Yang et al. \cite{Yang2019} & 2019 & \checkmark & \checkmark &  &  &  &  &  \\ 
\hline
John et al. \cite{John2019} & 2019 &  &  &  &  &  & \checkmark &  \\ 
\hline
Fahim et al. \cite{Fahim2020} & 2020 &  & \checkmark &  &  &  &  &  \\ 
\hline
Lyu et al. \cite{Lyu2020} & 2020 & \checkmark &  &  &  &  &  &  \\ 
\hline
Estabsari and Rajabi \cite{Estebsari2020} & 2020 & \checkmark & \checkmark & \checkmark &  &  &  &  \\ 
\hline
Ferraro et al. \cite{Ferraro2020} & 2020 & \checkmark &  &  &  &  &  &  \\ 
\hline
Xu et al. \cite{Xu2020b} & 2020 & \checkmark &  &  &  &  &  &  \\ 
\hline
Sreenivas et al. \cite{VandithSreenivas2021} & 2021 & \checkmark & \checkmark &  &  &  &  &  \\ 
\hline
Zhu et al. \cite{Zhu2021} & 2021 & \checkmark &  &  &  &  &  &  \\ 
\hline
Anjana et al. \cite{Anjana2021} & 2021 &  &  &  & \checkmark & \checkmark & \checkmark &  \\ 
\hline

Zhou and Kan \cite{Zhou2021} & 2021 & \checkmark &  &  &  &  &  &  \\ 
\hline
Sharma et al. \cite{Sharma2021} & 2021 & \checkmark &  &  &  &  &  &  \\ 
\hline
Chen et al. \cite{ChenandDeng2021} & 2021 &  &  &  & \checkmark &  &  &  \\ 
\hline
Jiang and Yen \cite{JiangandYen2021} & 2021 &  & \checkmark &  &  &  &  &  \\ 
\hline
Garcia et al. \cite{Garcia2020} & 2021 & \checkmark & \checkmark & \checkmark & \checkmark & \checkmark &  & \checkmark \\ 
\hline
Huang et al.\cite{Huang2023}& 2021 & \checkmark &  &  &  &  &  &  \\ 
\hline
Jiang et al. \cite{JiangandZhang2021} & 2021 & \checkmark & \checkmark &  &  &  &  &  \\ 
\hline
Singh et al. \cite{Singh2021} & 2021 &  &  &  & \checkmark &  &  &  \\ 
\hline
Santo et al. \cite{DeSanto2022} & 2022 & \checkmark & \checkmark & \checkmark &  & \checkmark &  &  \\ 
\hline
Chen and Wang  \cite{ChenandWang2022}& 2022 & \checkmark &  &  &  &  &  &  \\ 
\hline
Bertalanic et al. \cite{Bertalanic2022} & 2022 & \checkmark &  & \checkmark &  &  &  &  \\ 
\hline
Alsalemi et al. \cite{Alsalemi2022} & 2022 & \checkmark &  &  &  &  &  &  \\ 
\hline
Zhang et al. \cite{ZhangandHang2022} & 2022 &  &  & \checkmark & \checkmark &  &  &  \\ 
\hline
Dou et al. \cite{Dou2022} & 2022 &  &  &  &  & \checkmark &  &  \\ 
\hline
Abdel-Basse et al. \cite{Abdel-Basset2022}& 2022 & \checkmark & \checkmark & \checkmark &  &  &  &  \\ 
\hline
Wang et al. \cite{Wang2022} & 2022 & \checkmark & \checkmark &  & \checkmark &  &  & \checkmark \\ 
\hline
Bai et al. \cite{Bai2022} & 2022 & \checkmark & \checkmark &  &  &  &  &  \\ 
\hline
Abidi et al. \cite{Abidi2023} & 2023 & \checkmark & \checkmark & \checkmark &  &  &  &  \\ 
\hline
Paula et al. \cite{DePaula2023} & 2023 & \checkmark & \checkmark & \checkmark &  &  &  &  \\ 
\hline
Quan et al. \cite{Quan2023} & 2023 & \checkmark & \checkmark & \checkmark &  &  &  &  \\ 
\hline
Zhang et al. \cite{ZhangandShang2023} & 2023 & \checkmark &  &  &  &  &  &  \\ 
\hline
Copiaco et al. \cite{Copiaco2023} & 2023 &  &  &  &  &  &  & \checkmark \\ 
\hline
Qu et al. \cite{Qu2023} & 2023 & \checkmark & \checkmark &  &  &  &  & \checkmark \\ 
\hline
Sun et al. \cite{Sun2023}& 2023 & \checkmark & \checkmark &  &  &  &  &  \\ 
\hline
Sayed et al. \cite{Sayed2023} & 2023 &  &  &  &  &  &  & \checkmark \\ 
\hline
Hasan et al. \cite{Hasan2023} & 2023 & \checkmark &  &  &  &  &  &  \\
\hline
\end{tabular}
}
\end{table}


\subsection{Gramian Angular Field (GAF)}

Given a time-series is $X=\{x_1,x_2,...,x_N\}$, including $N$ samples, there are three steps to encode time-series into images \cite{Wang2015}. Firstly, $X$ time-series are scaled in the interval [0,1] according to Eq. (\ref{eq:1}).

\begin{equation} \label{eq:1}
 \widetilde{x_i} = \frac{\ x_{i}  - min(X)}{\max(X)-min(X)} 
\end{equation}

Then, the 1D time-series cartesian coordinate system is transformed into a polar coordinate system, which is a new representation of time-series. Angular cosine ($\phi$) and radius (r) are calculated to represent re-scaled time-series as polar coordinates using Eq. (\ref{eq:2}).

\begin{equation} \label{eq:2}
\left\{\begin{array}{c}
\phi=\arccos \left(\tilde{x}_i\right),-1 \leq \tilde{x}_i \leq 1, \tilde{x_i} \in \tilde{X} \\
r=\frac{t_{i}}{N}, t_{i} \in \mathbb{N}
\end{array}\right.
\end{equation}
where $t_i$ is the time stamp, and $N$ is a constant factor to regularize the span of the polar coordinate system.

There are two types of GAF based on the sum/difference of the trigonometric function, which are the Gramian Angular Summation Field (GASF) and the Gramian Angular Difference Field (GADF). GASF is defined in Eq. (\ref{eq:3}) and (\ref{eq:4}) and GADF is defined in Eq. (\ref{eq:5}) and Eq. (\ref{eq:6}).

\begin{equation}\label{eq:3}
GASF=\left[\begin{array}{ccc}
cos(\phi_{1}+\phi_{1}) & \cdots &cos(\phi_{1}+\phi_{n})  \\
cos(\phi_{2}+\phi_{1}) & \cdots &cos (\phi_{2}+\phi_{n})  \\
\vdots & \ddots & \vdots \\
cos(\phi_{n}+\phi_{1}) & \cdots &cos(\phi_{n}+\phi_{n}) 
\end{array}\right]
\end{equation}

\begin{equation}\label{eq:4}
GASF=  \widetilde{X} ' .\widetilde{X} -  \sqrt{I-  \widetilde{X} ^{2} }' . \sqrt{I-  \widetilde{X} ^{2}} 
\end{equation}

\begin{equation}\label{eq:5}
GADF=\left[\begin{array}{ccc}
sin(\phi_{1}+\phi_{1}) & \cdots &sin(\phi_{1}+\phi_{n})  \\
sin(\phi_{2}+\phi_{1}) & \cdots &sin (\phi_{2}+\phi_{n})  \\
\vdots & \ddots & \vdots \\
sin(\phi_{n}+\phi_{1}) & \cdots &sin(\phi_{n}+\phi_{n}) 
\end{array}\right]
\end{equation}

\begin{equation}\label{eq:6}
GADF=  -{\sqrt{I-  \widetilde{X} ^{2} }'}.\widetilde{X} -  \widetilde{X} '  .\sqrt{I-  \widetilde{X} ^{2} }
\end{equation}

In the above equations, $I$ refers to the unit vector [1,1,...,1].
\subsection{Markov Transition Fields (MTF)}
MTF is a powerful tool that keeps time domain information in time-series data by representing the sequential Markov transition probabilities. By utilizing the Markov matrix of quantile bins, MFT offers an approach to converting the time-series data into images \cite{Wang2015}. 

The encoding process begins a time-series $X$ and determines its $Q$ quantile bins. Secondly, each $x_i$ is assigned to its corresponding bin $q_j(j \in [1, Q])$. Then, the adjacency matrix $W=Q x Q$ is obtained, where each element  $w_{i,j}$ stands for the frequency at which a point in $q_j$ is followed by a point in $q_i$.Lastly, the Markov transition matrix is constructed as shown in Eq. (\ref{eq:7}).

\begin{equation}\label{eq:7}
W=\left[\begin{array}{ccc}
w_{i j \mid x_1 \in q_i, x_1 \in q_j} & \cdots & w_{i j \mid x_1 \in q_i, x_n \in q_j} \\
w_{i j \mid x_2 \in q_i, x_1 \in q_j} & \cdots & w_{i j \mid x_2 \in q_i, x_n \in q_j} \\
\vdots & \ddots & \vdots \\
w_{i j \mid x_n \in q_i, x_1 \in q_j} & \cdots & w_{i j \mid x_n \in q_i, x_n \in q_j}
\end{array}\right]
\end{equation}

\subsection{Reccurance Plot (RP)}

RP is a widely used tool to visualize and analyze the recurrent behaviors of time-series produced in dynamic framework \cite{Eckmann1987}. It is determined by a recursive matrix by computing the pairwise distance between the trajectories, in which the elements are calculated by the Eq. (\ref{eq:8}):

\begin{equation}\label{eq:8}
    R_{i,j} = \Theta ( \varepsilon - \mid  \mid   \overrightarrow{x} _{i} -\overrightarrow{x} _{j} \mid  \mid ) ,  i,j=1, \dots ,N
\end{equation}

Where $\epsilon$ is a threshold,  $\Theta$ is the Heaviside function used to binarize the distance matrixes, where its value is zero for the negative argument and one for the positive argument.

RP exposes the local correlation information of a sequence and hidden patterns by computing the distance matrix between subsequences.

\subsection{Short Time Fourier Transform (STFT) }

STFT can be considered as the frequency domain representation of the original signal. It utilized a window function to extract a part of the time domain signal and then performed a Fourier transform on it to specify diverse signal properties \cite{Gr_chenig_2001}. The STFT of a given signal y(x) is calculated in Eq. (\ref{eq:9}):

\begin{equation}\label{eq:9}
    STFT(n, \omega)= \sum_{- \infty }^\infty {y[ x ] \omega [n- x ] . e^{-j \omega n} }
\end{equation}

where $\omega(t)$ is the window function.

In addition. the spectrogram is generated by squaring the STFT magnitude as follows:

\begin{equation}\label{eq:10}
    Spectrogram(n, k)= \mid STFT(n,\omega)\mid^2
\end{equation}

\subsection{Continuous Wavelet Transformation (CWT)}

CWT offers an unstable window size that adjusts based on the frequency at the cost of the time resolution. Although STFT provides a great representation of the signal's time-frequency characteristics, it presents a fixed resolution in the frequency domain, which is not always ideal in certain scenarios. On the other hand, CWT is an operation linearly on a time-domain signal y(t) given by:

\begin{equation} \label{eq:11}
W_{a,b}[y(t)] = \frac{1}{\sqrt{a}}\int^\infty_{-\infty} y(t)  \ast  \phi \left(\frac{t-b}{a}\right) {dt}
\end{equation}

where $\phi(\frac{t-b}{a})$ is a dilated version of the base wavelet function $\phi(t)$ by applying scaling and shifting. $a > 0$ is the scaling variable that regulates the spread of the function, and $b$ is the time-shifting parameter or the instant of time at which the signal needs to be analyzed.
The visual representation of the CWT of a signal is referred to as a scalogram \cite{Bol_s_2014}.

\subsection{Hilbert Huang Transform (HHT)}

HHT is an analysis of signals that is non-stationary and non-linear \cite{Huang1998}. While most techniques may fail in analyzing nonstationary and nonlinear systems, HHT alleviates the challenges of time-frequency-energy representation of the data. HHT includes two primary phases, called Empirical Mode Decomposition (EMD) and Hilbert Transform (HT). The transformation involves several processes. First, EMD is utilized to obtain Intrinsic Mode Functions (IMFs) from the signals. Second, the Hilbert transform is applied to each of the IMF components. Finally, the instant frequency and amplitude can be computed.

\subsection{Other Transformation Methods}

In addition to the aforementioned methods, the literature offers a range of alternative techniques that are commonly employed to address various types of problems. These methods play a pivotal role in the transformation of IoT time-series data. Some of the notable approaches in this regard include data normalization combined with matrix conversion, the direct drawing method, Gaussian Mixture Regression (GMR), Gray-Scale encoding (GS), Gray-Scale image representations, RGB color image conversion, the Wavelet Variance Image (WVI) method. These techniques have gained popularity within the literature for their effectiveness in transforming and enhancing the analysis of IoT time-series data.

Garcia et al. \cite{Garcia2020} proposed a modification of GS by choosing lower and upper bounds in original formulations in accordance with the GAF encoding instead of minimum and maximum scaling. Wang et al. \cite{Wang2022} used the direct drawing method which signals are transformed into a 2D spectrum map directly with plt functions in the Matplotlib package in Python without any processing. The direct drawing method has higher accuracy than GAF and MTF after STFT. The main idea of GS is to transform time-domain raw signals into images. The time-domain raw signals complete the pixels of the image sequentially. Wen et al. \cite{Wen2018} reorganized the GS using CNN for fault diagnosis in manufacturing systems. A transformation method consisting of data normalization and matrix conversion was used for 2D image representation \cite{Sayed2023}, \cite{Copiaco2023}.  1D time-series data is first normalized in [0,1] with n features. Then, these features are arranged in $mxm$ matrix format. Lastly, this matrix is resized to 28x28 pixels and saved as an image to obtain a gray-scale image or RGB color image. The Voltage–Current (VI) trajectory can be converted into a pixelated VI image ($nxn$ matrix) by meshing the VI trajectory \cite{DeBaets2018}. Qu et al. \cite{Qu2023} generated 2D load signatures according to the corresponding features of the signal based on the Weighted Voltage–Current (WVI) trajectory image.


\begin{table*}
\centering
\begin{threeparttable}
\caption{Summary of Image Transformation Application According to Data Types}
\label{table:SurveyonDataTypes}
\begin{tabular}{|>{\centering\hspace{0pt}}m{0.025\linewidth}|>{\centering\hspace{0pt}}m{0.07\linewidth}|>{\centering\hspace{0pt}}m{0.07\linewidth}|>{\hspace{0pt}}m{0.235\linewidth}|>{\centering\hspace{0pt}}m{0.055\linewidth}|>{\centering\hspace{0pt}}m{0.080\linewidth}|>{\hspace{0pt}}m{0.30\linewidth}|} 
\hline
\textbf{Ref.} & \textbf{Univariate} & \textbf{Multivariate} & \multicolumn{1}{>{\centering\hspace{0pt}}m{0.235\linewidth}|}{\textbf{Dataset}} & \textbf{Gray / Colored} & \textbf{Methods} & \multicolumn{1}{>{\centering\arraybackslash\hspace{0pt}}m{0.30\linewidth}|}{\textbf{Fusion Techniques}} 
 \\ \hline

\cite{ZhangandShang2023}& \checkmark &  & -Case Western Reserve University (CWRU)\par{}-Autonomous Experimental dataset & Color\tnote{*}  & GASF\par{}GADF & - \\
\hline
 \cite{Baldini2019} & \checkmark &  & - Private dataset of RF emissions collected from 11 IoT devices & Gray & RP & - \\
\hline
\cite{Lyu2020} & \checkmark & & - Private dataset which collected fiber intrusion disturbance signals & Color\tnote{*}
 & GAF & - \\* 
\hline
 \cite{Zhu2021}& \checkmark & & - KDD Cup 99 data & Color\tnote{*}
 & GAF & - \\* 
\hline
 \cite{Bertalanic2022}& \checkmark &  & - Rutgers dataset & Gray\par{}Color
 & RP\par{}GAF & - \\* 
\hline
\cite{Fahim2020}& \checkmark &  & - REFIT electrical load measurement dataset & Color\tnote{*} & MTF & - \\* 
\hline
 \cite{Estebsari2020}& \checkmark & & - Boston housing price data\par{}- Load Forecasting Dataset & Color\tnote{*} & RP\par{}GAF\par{}MTF & - \\* 
\hline
 \cite{John2019}& \checkmark & & -Private dataset\par{}-Physionet/ Computing in Cardiology (CinC) Challenge 2016 & Color\tnote{*} & HHT & - \\* 
\hline
\cite{Anjana2021} & \checkmark & & - Seed & Color\tnote{*} & STFT\par{}CWT\par{}HHT & - \\* 
\hline
   \cite{ChenandDeng2021}& \checkmark &  & - Private dataset & Color\tnote{*} & STFT & - \\* 
\hline
  \cite{Singh2021}& \checkmark & & - TUH Abnormal EEG Corpus & Color\tnote{*} & STFT & - \\* 
\hline
   \cite{ZhangandHang2022}& \checkmark &  & - Arrhythmia Data\par{}- Private dataset & Gray\par{} Color & RP\par{}STFT & - \\* 
\hline
   \cite{Dou2022}& \checkmark & & - MIT-BIH arrhythmia\par{}- MIT-BIH normal sinus rhythm\par{}- BIDMC & Color\tnote{*}
 &CWT & - \\* 
\hline
   \cite{DePaula2023} & \checkmark & & - Private dataset &  Gray\par{} Color & GADF\par{}GASF\par{}MTF\par{}RP &  \\* 
\hline
 \cite{JiangandYen2021} & \checkmark && - Private dataset & Color\tnote{*}
 & MTF & - \\* 
\hline
   \cite{Garcia2020} & \checkmark && - Airbus SAS Airbus SAS 2018 & Gray\par{}Color & GAF\par{}MTF\par{}RP\par{}GS\par{}STFT\par{}DWT & - \\* 
\hline
 \cite{Bai2022} & \checkmark && - Private dataset & Color & GAF\par{}MTF & - \\* 
\hline
 \cite{Hasan2023}& \checkmark && - WSN Dataset\par{}- ETDataset\par{}- TON\_IoT Dataset &  Color\tnote{*} & GAF & - \\* 
\hline
   \cite{VandithSreenivas2021}& \checkmark & & - MIT-BIH arrhythmia database & Color  & GAF\par{}MTF & - \\* 
\hline
   \cite{Alsalemi2022}& \checkmark &&  &  Color & GAF & - \\* 
\hline
 \cite{Huang2023} & \checkmark & & - Caltrans Performance Management System (PeMS) & Color\tnote{*} & GASF & - \\* 
\hline
  \cite{Xu2020b}& \checkmark & & \par{}- WISDM\par{}- UCI HAR\par{}- OPPORTUNITY & Color & GASF\par{}GADF & - \\* 
\hline
\cite{Zhou2021} &  & \checkmark & \par{}- 2018 China Physiological Signal Challenge (CPSC2018)\par{}- PhysioNet Long-term ST dataset & Color\tnote{*} & GADF & - Each channel of the ECG signal transform into a GAF image which is represented as a 2nd-order ECG tensor.\par{}- These images are then stacked together to form a 3rd-order ECG tensor by concatenating them along the 3rd dimension.

 \\* 
\hline
   \cite{ChenandWang2022} &  & \checkmark & - PLAID \par{}- WHITED & Color & GAF & - Single-channel images correspond to three channels in the RGB Color space, respectively, to create an RGB image.
 \\* 
\hline
 \cite{Copiaco2023}&  & \checkmark  & - The Simulated Energy Dataset (SiD)\par{}- The Dutch Residential Energy Dataset (DRED) & Gray\par{} Color  & a grayscale image\par{}an RGB color image\par{}(jet colormap) & - A 5 × 5 matrix is used to organize features for a given instant.
 
 \par{}- Then, the matrix is resized to 28x28 pixels and saved as a Grayscale or an RGB Color image.
 \\* 
\hline

\end{tabular}

  \end{threeparttable}
\end{table*}

\begin{table*}
\centering
\begin{threeparttable}

\caption* {\textbf{Table \ref{table:SurveyonDataTypes} Continued:}  Summary of Image Transformation Application According to Data Types}

\begin{tabular}{|>{\centering\hspace{0pt}}m{0.025\linewidth}|>{\centering\hspace{0pt}}m{0.07\linewidth}|>{\centering\hspace{0pt}}m{0.07\linewidth}|>{\hspace{0pt}}m{0.235\linewidth}|>{\centering\hspace{0pt}}m{0.055\linewidth}|>{\centering\hspace{0pt}}m{0.080\linewidth}|>{\hspace{0pt}}m{0.30\linewidth}|} 
\hline
\textbf{Ref.} & \textbf{Univariate} & \textbf{Multivariate} & \multicolumn{1}{>{\centering\hspace{0pt}}m{0.235\linewidth}|}{\textbf{Dataset}} & \textbf{Gray / Colored} & \textbf{Methods} & \multicolumn{1}{>{\centering\arraybackslash\hspace{0pt}}m{0.30\linewidth}|}{\textbf{Fusion Techniques}} 
 \\ \hline

 \cite{Qu2023}&  & \checkmark  & - PLAID\par{}- WHITED\par{}- HRAD & Color\tnote{*} & MTF\par{}GAF\par{}WVI & - Each variable convert into images using three encoding techniques. \par{}
 - Then, the WVI image and MTF image are superimposed to create two channels. Also, the I-GAF image is saved as a new image by the Energy-Normalization (EN) block. \par{}
 - Lastly,  this image is superimposed with the other two images to get a three-channel image.
 \\* 

\hline

  \cite{Sharma2021}&  & \checkmark & - 1D Biomedical Signals such as ECG, PPG, temperature, and accelerometer & Color & GAF & -The average of the computed features from various channels is found and provided a single fused feature set using Channel-Wise Mean Fusion (CAF).
 \\* 
\hline
 \cite{Abdel-Basset2022} &  & \checkmark & - UCI HHAR \par{}- UCI MEHEALTH & Color & RP\par{}MTF\par{}GAF & - Each row includes three measures as x, y, and z for AM, GY and MG data in 3D, respectively. \par{}- Converts x-, y- and z-axis of signals as red, green, and blue channels of images.
 \\* 
\hline
   \cite{Ferraro2020}&  & \checkmark & - Backblaze SMART dataset & Color\tnote{*} & GAF & - Each feature of time series is transforming into polar coordinate through the GAF.
 \\* 
\hline
   \cite{DeSanto2022} &  & \checkmark & - PAKDD2020 Alibaba AI OPS Competition\par{}- NASA bearings & Color\tnote{*} & RP\par{}GAF\par{}MTF\par{}STFT & - MTS encoded a set of feature maps which computing with four different image transformation techniques.
 \\* 
\hline
 \cite{Sun2023}&  & \checkmark & - Private dataset & Color & GAF\par{}MTF & - GASF, GADF, and MTF layer are placed on the red, green, and blue layer, respectively and saved images.
 \\* 
\hline
 \cite{Abidi2023}&  & \checkmark  & - SITS data which collected for a different study of Dordogne\par{}- Reunion Island study\par{}- Koumbia &Color\tnote{*}  & \par{}GADF\par{}GASF\par{}MTF\par{}RP & - Each UTS in MTS is flattened to the direct use of MTS instead of thinking independently of each UTS. \par{}- Then, generate the 2D images from the flattened MTS.
 \\* 
\hline
    \cite{Sayed2023}&  & \checkmark & - Student Room Dataset (SRD)\par{}- UCI dataset (an office space)\par{}- Living Room Dataset (LRD) & Gray\par{} Color  & data normalization \par{}matrix conversion & - The list of features(n) of the dataset is arranged into 3xn matrix format. \par{}- Then, The matrix is resized to 28x28 pixel and saved as an image.
 \\* 
\hline
 \cite{Yang2021} &  & \checkmark  & - Wafer dataset\par{}- ECG dataset & Color & GASF\par{}GADF\par{}MTF & - Encode MTS as Colored image for each univariate time-series. Each Colored image is separated into three monochronic images, namely red, green, and blue (RGB). \par{}- After the separation, these mono-color images are concatenated together to form a huge image.
 \\* 
\hline
\cite{Wang2022} & \checkmark & \checkmark  & - Case Western Reserve University (CWRU) Dataset\par{}- Society for Machinery Failure Prevention Technology (MFPT) Dataset & Color\tnote{*} & STFT\par{}The direct drawing method\par{} GADF \par{}MTF & - The vibration signals from multiple channels are combined into a 2D spectrum map.
 \\* 
\hline
   \cite{Quan2023}& \checkmark & \checkmark & - Chinatown\par{}+84 UCR datasets & Color\tnote{*} & GAF\par{}MTF\par{}RP\par{}GMR & - 1D multi-scale features and 2D image features are fused in two distinct methods, covering the feature fusion methods such as SE and SA \par{}-Three images which are encoded with different coding methods are overlapped as three-channel data inputs.
 \\* 
\hline
   \cite{JiangandZhang2021} & \checkmark & \checkmark & - 24 benchmark datasets\par{}(14 dataset for MTS and \par{}10 dataset for UTS) & Color \tnote{*} & GAF\par{}MTF & - Each UTS in MTS are converted into GM images. Each variable is considered a channel.
\par{}- G-image and M-image are concatenated  GM-feature maps by the adaptive feature aggregation, which pass through a corresponding shallow CNN separately. \\
\hline
\end{tabular}
\begin{tablenotes}
      \item[*] {The color is not specified in the paper. For this reason, the color is determined based on the given images.}
    \end{tablenotes}
  \end{threeparttable}
\end{table*}

\section{Image Transformation in IoT Applications}

IoT encompasses various domains where time-series data is frequently used. Time-series data is a data type that includes a sequence of data points that are collected at regular intervals over time. Table ~\ref{table:SummaryStudies} summarizes the existing studies by categorizing them according to nine IoT domains. Here are some IoT domains where time-series data is commonly utilized:

\subsection{Security and Privacy}

The security and privacy domain within IoT focuses on handling the challenges and risks associated with ensuring the confidentiality, integrity, availability, and privacy of IoT systems, devices, and data. 
Security in the IoT domain includes implementation preventions to obstruct unauthorized access, data breaches, and malicious activities that have the potential to jeopardize the functionality, integrity, and confidentiality of IoT devices and systems. 
Privacy in the IoT domain refers to the protection of individual's personal information and their control over how it is collected, used, and shared by IoT systems.

IoT time-series data plays a significant role in the security and privacy domain by providing valuable insights into the behavior, patterns, and anomalies within IoT systems. Anonymization, encryption, and access controls should be applied appropriately to protect sensitive information contained within the time-series data. In the context of IoT security and privacy, time-series data can be leveraged for various purposes: Intrusion detection,  unauthorized access detection, anomaly detection,  security analytics prediction, etc.

Baldini et al. \cite{Baldini2019} presented an approach for the authentication of IoT wireless devices based on Radio Frequency (RF) emissions. The proposed approach combined CNN and RP (RP-CNN) is tested on the RF emissions dataset, which is experimental data collected from 11 IoT devices. They also applied two classification methods called T-CNN which utilizes the digital representation of the RF emissions directly with CNN, and FEAT, which extracted the statistical characteristics of RF emissions from their digital representations. The results showed that the RP-CNN improves accuracy when compared to T-CNN and FEAT. 
Lyu et al. \cite{Lyu2020} proposed an intrusion pattern recognition framework. The method, based on the GAF and CNN, achieved a high-speed response time of 0.58 s and a high recognition accuracy of 97.57\% for six types of optical fiber intrusion events. In addition, it improved the robustness and practicability of the system because the GAF algorithm is not sensitive to the fluctuation of power sources in the optical path.
Zhu et al.\cite{Zhu2021} developed a monitoring system to detect abnormal traffic and vulnerability attacks in IoT applications. In the system, time-series data was converted into GAF graphs, and the CNN and Long Short-term Memory (LSTM) combination model was utilized to monitor traffic. However, the system that combined C5.0 decision tree and time-series analysis introduced a novel idea for the traffic analysis of IoT devices.
Bertalani{\v{c}} et al. \cite{Bertalanic2022} proposed a new resource-aware approach based on image transformation and deep learning for anomaly detection in the wireless link layer. time-series data were transformed into images using RP and GAF. The experiments show that RP outperforms the GAF methods by up to \%14.

\subsection{Energy Management}

IoT enables the monitoring and control of energy consumption, smart grid management, and the integration of renewable energy sources. It helps optimize energy distribution, reduce waste, and improve sustainability.

Fahim et al. \cite{Fahim2020} proposed a model called Time-series to Image (TSI) to detect abnormal energy consumption in residential buildings. This study focused on analyzing the univariate time-series energy data for very short-term analysis. The Proposed model utilized a One-Class Support Vector Machine (OCSVM) as a classifier and MTF as a converter, which transforms univariate time-series data into images. In this work, the authors demonstrated that this image representation further enhances the classifier's ability to detect anomalous behavior more efficiently. Estebsari and Rajabi \cite{Estebsari2020} proposed a hybrid model based on CNN and image encoding methods for single residential loads. They applied three different image encoding methods, including the RP, GAF, and MTF, to historical load time-series data. The experiments revealed that RP performed the best among the three encoding methods. Alsalemi et al. \cite{Alsalemi2022} developed a novel GAF classifier based on the EfficientNet-B0 for the classification of edge internet of energy applications. The authors aimed to introduce the first lightweight classifier for 2D energy consumption working on the ODROID-XU4 platform.

Copiaco et al. \cite{Copiaco2023} proposed a 2D pre-trained CNN model for detecting anomalies in building energy consumption. This model used the 2D versions of the energy time series signals to give input to several pre-trained models, such as AlexNet and GoogleNet, as features of the Linear Support Vector Machine (SVM) classifier. In this study, 1D time-series were transformed into Grayscale and Jet Color image representations. This study showed that converting energy time-series data into images can provide an increment of the correlation between images with the same class.
Chen and Wang \cite{ChenandWang2022} proposed an edge-computing architecture for load recognition tasks in the field of Non-Intrusive Load Monitoring (NILM) that reduces data transmission volume and network bandwidth requirements. They also developed a color encoding method based on GAF to construct load signatures in home appliances.
Qu al. \cite{Qu2023} constructed three 2D load signatures based on the WVI, MTF, and current spectral sequence-based GAF (I-GAF). Additionally, they designed a new Residual Convolutional Neural Network with Squeeze-and-Excitation (SE) and Energy-Normalization (EN) blocks (EN-SE-RECNN) for appliance identification in NILM. This study compared the performance of various models, including Residual Convolutional Neural Network (RECNN), Residual Convolutional Neural Network with EN blocks (EN-RECNN), and EN-SE-RECNN, and confirmed that the performance of EN-SE-RECNN was better. Also, their findings demonstrate that the fusion of different signatures enhances performance by enriching the information related to appliance identification.

\subsection{Healthcare}

In healthcare applications, time-series data assists in monitoring patient vital signs, analyzing health trends, predicting disease outbreaks, and optimizing healthcare resource allocation. Zhou and Kan \cite{Zhou2021} developed a tensor-based framework for ECG anomaly detection in Internet of Health Things (IoHT)-based cardiac monitoring and smart management of cardiac health. The multi-channel ECG signals were converted into 2D images using GADF. The proposed model demonstrated that the framework using 2D image representations shows better performances than that directly using 1D signals because of the difficulty of extracting hidden information.

Sreenivas et al. \cite{VandithSreenivas2021} proposed a CNN model for the classification of arrhythmia in dual-channel ECG signals. In this study, GAF and MTF were used to convert the ECG time-series signals into images. The result showed that the GAF model achieved higher accuracy compared to the MTF. Anjana et al. \cite{Anjana2021} proposed a CNN model based on various types of image encoding approaches to classify human emotions using EEG signals. In this study, Spectrogram, Scalogram, and HHT were employed to transform EEG signal data into images. The experiments showed that the scalogram of image encoding provides the best classification accuracy. Paula et al. \cite{DePaula2023} proposed a 2D-kernel-based CNN architecture to classify the Steady State Visually Evoked Potentials (SSVEP) signal. In this work, EEG data is encoded into images using GADF, GASF, MTF, and RP. This study demonstrated that the GADF and RP methods consistently showed higher performance. Also, the 1D-kernel-based structure of the model was insufficient for learning the necessary information from the data.

John et al. \cite{John2019} developed a cardiac monitoring system based on wireless sensing, aiming for accurate diagnosis of heart diseases. The system used MQTT for long-distance transmission and HTT for preprocessing and feature extraction of the data. Sharma et al. \cite{Sharma2021} introduced a patient monitoring system based on ontology for early remote detection of COVID-19. The proposed system relied on an alarm-enabled bio-wearable sensor system that utilized sensory 1D biomedical signals such as ECG, PPG, temperature, and accelerometer. These 1D  Biomedical signals were converted into images with GASF after extracting their features. Then, SVM and K-Nearest Neighbors (KNN) were employed as ML-based classifiers for the classification of COVID-19 patients.
Chen et al.\cite{ChenandDeng2021} proposed an indoor speed estimation framework, SpeedNet, from radio signals, mainly aimed at monitoring the movement of elderly individuals. The SpeedNet framework includes three modules: the dominant path extraction module, the spectrum analysis module, and the deep learning module. The dominant path signal which is obtained from the extraction module was analyzed using STFT in the spectrum analysis module. Also, CNN and LSTM were utilized in the deep learning module to extract spatial and temporal features. They introduced a new approach for contactless indoor speed estimation with radio signals, addressing the challenges posed by the complex relationship between the speed of moving individuals and radio signals.
Singh et al. \cite{Singh2021} suggested a brain signal classification model that transformed brain signals into images as input for a pre-trained VGG19 nıodel by using STFT for seizure detection. In addition, blockchain technology was utilized to store images more securely. The study also emphasizes the importance of selecting an appropriate encoding method, which involves using different image conversion techniques such as spectrograms, chronograms, or kurtograms.
Zhang et al. \cite{ZhangandHang2022} proposed a system based on 5G-enabled Medical IoT for automatic detection of arrhythmia (ARR). Time-frequency spectrograms obtained from RR interval sequences using RP and Fourier Transform (FT) were used as inputs to a unified CNN and LSTM model for the classification of ECG signals. Dou et al. \cite{Dou2022} proposed a novel classification method based on CWT and CNN within the context of IoT domain. Their approach simultaneously classifies various ECG signals for heart disease diagnosis using GoogleNet. Besides, ECG signals were converted into time-frequency images with CWT. Abdel-Basset et al. \cite{Abdel-Basset2022} developed a lightweight Human Activity Recognition (HAR) architecture designed to classify human activities captured by heterogeneous sensors from different IoT devices. They proposed a few modifications for three encoding techniques, including RP, MTF, and GAF. These techniques encode the three-dimensional (3D) time-series data of human activities into three-channel images to overcome the heterogeneity in sensory data.

\subsection{Industrial}

Industrial IoT (IIoT) involves connecting industrial equipment, machinery, and systems to enable data monitoring, analysis, and optimization in manufacturing, energy transportation, and other industrial sectors. In industrial settings, time-series data helps monitor equipment performance, predict failures, optimize maintenance schedules, and improve overall operational efficiency.

Various image encoding methods are commonly used in IIoT to provide intelligent and efficient fault diagnosis. Wang et al. \cite{Wang2022} proposed a framework for fault diagnosis of single-channel and multi-channel bearing signals. They combined spectrum map information fusion and CNN to achieve fast fault diagnosis. GADF, MTF, and STFT were used to generate a 2D spectrum graph from 1D bearing vibration data, and STFT achieved the best result with the lowest loss value. The experiments indicated that the STFT method could use multichannel information effectively and improve fault identification accuracy. Similarly, Zhang et al. \cite{ZhangandShang2023} presented a novel fault diagnosis method that combines GAF, Extreme Learning Machine (ELM), and CNN. They explored different encoding methods, including GADF, GASF, spectrogram, and gray-scale image, to indicate the effectiveness of the chosen encoding techniques for pattern recognition. The findings indicated that the GADF has the highest performance. Santo et al. \cite{DeSanto2022} developed a model that combined time-series encoding techniques and CNN  for predictive maintenance. Four main encoding techniques, such as RP, GAF, MTF, and Wavelet transform, were evaluated in this paper.  The RP achieved the best performance in all metrics.

Ferraro et al. \cite{Ferraro2020} developed an efficient method for predictive maintenance that improved maintenance strategies and decreased downtime and cost. The method involves transforming temporal time-series data into images using GAF and utilizing deep learning strategies to predict the health status of the Hard Disk Drive (HDD). Jiang et al. \cite{JiangandYen2021} proposed the MTF-CLSTM method, which combines the MTF, CNN, and LSTM to predict product quality in Wire Electrical Discharge Machining (WEDM). MTF is employed to transform dynamic WEDM manufacturing conditions into images. In addition, Featurtes were extracted from the images with CNN, and LSTM is used to predict the surface roughness of the WEDM products right after manufacturing. When MTF-CLSTM method was compared with the Deep Neural Network (DNN) and the Markov Chain DNN(MC-DNN) methods \cite{Fan2019}, the proposed method achieved the best performance.

Garcia et al. \cite{Garcia2020} explored six encoding methods (GAF, MTF, RP, GS, spectrogram, and scalogram) and the modifications to enhance their robustness against the variability in large datasets when transforming temporal signals into images. This study revealed that different encoding methods exhibit competitive results for anomaly detection in large datasets.
Bai et al. \cite{Bai2022} proposed a fault diagnosis method called Time-series Conversion-DCGAN (TSC-DCGAN). They utilized GAF and MTF for transforming 1D electrical parameters into 2D images. Additionally, the Deep Convolutional Generative Adversarial Network (DCGAN) was used as a generation method to handle the inadequate data samples of electrical parameters from oil wells. Also, The experimental results show that GAF images performed better in terms of classification effectiveness compared to MTF images. Sun et al. \cite{Sun2023} put forward an idea for diagnosing composite failures of the adaptable multi-sensor bearing gear system by leveraging GAF, MTF, and ResNet. The complicated multi-dimensional time-series signals were fused and transformed into 2D images to facilitate classification tasks using GAF and MTF.

\subsection{Environmental Monitoring}

IoT devices are exploited to monitor and manage environmental conditions such as air quality, water quality, pollution levels, and natural resource conservation. These solutions aid in environmental protection and sustainable practices.

Abidi et al. \cite{Abidi2023} proposed a framework for the classification of Land Use/Land Cover (LULC) mapping based on 2D encoded multivariate Satellite Image Time-series (SITS). In this work, multivariate SITS data was converted into 2D images by GADF, GASF, MTF, and RP. The results indicated that the RP technique performed better than all encoding techniques. In addition, the combination of 2D encoding techniques achieved better performance than using the encoding methods alone.

\subsection{Smart Building}
Smart Building enhances occupant comfort, reduces energy consumption, improves safety and security, and optimizes building operations and maintenance. time-series data in smart buildings is employed to monitor and control various building systems, such as HVAC (Heating, Ventilation, and Air Conditioning), lighting, and occupancy.

Sayed et al. \cite{Sayed2023} presented an approach for the detection of occupancy using environmental sensor data such as temperature, humidity, and light sensors. In this study, multivariate time-series data was transformed into gray-scale and RGB images using an image transformation method to encode better and obtain relevant features. This method covered data normalization and matrix conversion, unlike commonly used methods such as GAF. The results showed that gray-scale images offer to provide appropriate balance between accuracy and training time compared to the colored images.

\scriptsize

\renewcommand{\thefootnote}{\fnsymbol{footnote}}
\begin{table*}
\centering
\begin{threeparttable}
\caption{Summary of Image Transformation Techniques Studies in IoT Application Domain} \label{table:SummaryStudies}
\begin{tabular}{>{\hspace{0pt}}m{0.080\linewidth}>
{\centering\hspace{0pt}}m{0.031\linewidth}>
{\centering\hspace{0pt}}m{0.032\linewidth}>
{\centering\hspace{0pt}}m{0.093\linewidth}>
{\centering\hspace{0pt}}m{0.090\linewidth}>
{\centering\hspace{0pt}}m{0.059\linewidth}>
{\centering\hspace{0pt}}m{0.10\linewidth}>
{\centering\hspace{0pt}}m{0.08\linewidth}>
{\centering\hspace{0pt}}m{0.06\linewidth}>
{\centering\arraybackslash\hspace{0pt}}m{0.14\linewidth}} 

\\\hline
\textbf{Domain} & \textbf{Ref.} & \textbf{Year} & \textbf{Problem Type} & \textbf{Application Type} & \textbf{Methods} & \textbf{Models} & \textbf{Comparision Models} & \multicolumn{1}{>{\centering\hspace{0pt}}m{0.06\linewidth}}{\textbf{Performance}\par{}\textbf{ Metrics}} & \textbf{Results} \\* 
\hline
\multirow{4}{0.071\linewidth}{\hspace{0pt}\centering{}Security} & \cite{Baldini2019} & 2018 & Authentication & Authentication of IoT devices & RP & CNN & T-CNN\par{}FEAT & Classification Accuracy\par{}Confusion Matrix\par{}Accuracy & Accuracy:\par{}RP-CNN --\textgreater{} 96.8\%\par{}T-CNN --\textgreater{} 96.2\%\par{}FEAT --\textgreater{} 91.3\% \\* 
\cline{2-10}
 & \cite{Lyu2020} & 2020 & Classification & Intrusion Pattern Recognition & GAF & CNN & VGG16\par{}ResNet50\par{}Inception V3 & Precision\par{}Recall\par{}F-Score & Accuracy: \par{}97.67\% \\* 
\cline{2-10}
 &  \cite{Zhu2021} & 2021 & Classification & Anomaly Detection & GAF & C5.0 Decision Tree (DT)\par{}CNN-LSTM & - & Accuracy\par{}Detection Rate (DR)\par{}False Positive Rate (FAR) & Accuracy: \par{}96\% \\* 
\cline{2-10}
 &  \cite{Bertalanic2022} & 2022 & Classification & Anomaly Detection & RP\par{}GAF & CNN & KNN\par{}SVM\par{}AlexNet\par{}VGG11 & F1-score\par{}Precision\par{}Recall & F1-score:\par{}SuddenD: 1.00\par{}SuddenR: 1.00\par{}InstaD: 0.92\par{}SlowD: 0.99\par{}No anomaly: 0.99 \\* 
\hline
\multirow{6}{1.0cm}{\hspace{0pt}{}Energy Management} &  \cite{Fahim2020}& 2020 & Detection & Anomaly Detection & MTF & OCSVM & Principal Component Analysis (PCA)+OCSVM & F1-score\par{}Precision\par{}Recall & F1-score: \par{}88\% \\* 
\cline{2-10}
 &  \cite{Estebsari2020} & 2020 & Prediction & Single Residential\par{}Load Forecasting & RP\par{}GAF\par{}MTF & CNN & SVM\par{}Artificial Neural Network (ANN)\par{}1D-CNN & MAE\par{}MAPE\par{}RMSE & MAE: 0.59\par{}MAPE: 12.54\par{}RMSE: 0.79 \\* 
\cline{2-10}
 &  \cite{Alsalemi2022} & 2022 & Classification & Energy Consumption Data Classification & GAF & EfficientNet-B0 &-  & - & -  \\* 
\cline{2-10}
 & \cite{ChenandWang2022} & 2022 & Recognition & Load Recognition & GAF & ResNet & Other Ref. Papers (LSTM,CNN and three AlexNet versions) & Precision\par{}Recall\par{}F-Score\par{}Confusion Matrix & Accuracy: \par{}PLAID --\textgreater{}97.97\%\par{}WHITED --\textgreater{} 97.90\% \\* 
\cline{2-10}
 &  \cite{Copiaco2023} & 2023 & Detection & Anomaly Detection & \par{}Grayscale image\par{}RGB color image\par{}(jet colormap) & \par{}AlexNet\par{}GoogleNet\par{}SqueezeNet\par{}Linear SVM & AlexNet\par{}GoogleNet & F1-scores\par{}Specificity\par{}Precision\par{}Recall\par{}Accuracy & F1-scores: \par{}SiD --\textgreater{} 93.63\% \par{}DRED --\textgreater{} 99.89\% \par{}Accuracy: \par{}SiD --\textgreater{} 96.11 \par{}DRED --\textgreater{} 99.91\% \\* 
\cline{2-10}
 & \cite{Qu2023} & 2023 & Recognition & Load Recognition & MTF\par{}GAF\par{}WVI & EN-SE-RECNN & RECNN\par{}EN-RECNN\par{}EN-SE-RECNN & Accuracy & Accuracy:\par{}PLAID --\textgreater{} 97.43\%\par{}WHITED --\textgreater{} 95.99\%\par{}HRAD --\textgreater{} 98.14\% \\* 
\hline
\multirow{3}{0.071\linewidth}{\hspace{0pt}{}Healthcare} &  \cite{John2019}& 2019 & Classification & Cardiac Monitoring System & HHT & Adaptive Threshold Method & - & Accuracy & Accuracy: \par{}96\% \\* 
\cline{2-10}
 &  \cite{VandithSreenivas2021} & 2021 & Classification & Arrhythmia Classification & GAF\par{}MTF & CNN & Other Papers & Accuracy & Accuracy:\par{}GAF --\textgreater{} 97\%\par{}MTF --\textgreater{} 85\% \\* 
\cline{2-10}
 &  \cite{Zhou2021} & 2021 & Detection & Anomaly Detection & GADF & -Deep Support Vector Data Description (Deep SVDD)\par{}-Statistical Control Charts (e.g., Hotelling T2 chart)\par{}-MPCA & Adaboost \par{}SVM & Accuracy\par{}Area Under the ROC Curve (AUROC)\par{}F-score & F1-score: \par{}Atrial fibrillation -\textgreater{} 0.9771 \par{}Right bundle branch block -\textgreater{} 0.9986\par{}ST-depression -\textgreater{} 0.9550 \\* 
\cline{2-10}

\end{tabular}
  \end{threeparttable}
\end{table*}

\begin{table*}
\centering
\begin{threeparttable}
\caption* {\textbf{Table \ref{table:SummaryStudies} Continued:} Summary of Image Transformation Techniques Studies in IoT Application Domain} 
\begin{tabular}{>{\hspace{0pt}}m{0.080\linewidth}>
{\centering\hspace{0pt}}m{0.031\linewidth}>
{\centering\hspace{0pt}}m{0.032\linewidth}>
{\centering\hspace{0pt}}m{0.093\linewidth}>
{\centering\hspace{0pt}}m{0.090\linewidth}>
{\centering\hspace{0pt}}m{0.059\linewidth}>
{\centering\hspace{0pt}}m{0.10\linewidth}>
{\centering\hspace{0pt}}m{0.08\linewidth}>
{\centering\hspace{0pt}}m{0.06\linewidth}>
{\centering\arraybackslash\hspace{0pt}}m{0.14\linewidth}} 

\\\hline
\textbf{Domain} & \textbf{Ref.} & \textbf{Year} & \textbf{Problem Type} & \textbf{Application Type} & \textbf{Methods} & \textbf{Models} & \textbf{Comparision Models} & \multicolumn{1}{>{\centering\hspace{0pt}}m{0.06\linewidth}}{\textbf{Performance}\par{}\textbf{ Metrics}} & \textbf{Results} \\* 
\hline
\multirow{8}{0.071\linewidth}{\hspace{0pt}{}Healthcare} & 
 \cite{Anjana2021} & 2021 & Classification & Emotion Classification & STFT\par{}CWT\par{} HHT & CNN & - & Accuracy\par{}Precision\par{}Recall\par{}F1-score & Accuracy:\par{}Scalogram --\textgreater{} 98\%\par{}Spectrogram --\textgreater{} 78\%\par{} HHT --\textgreater{} 75\% \\* 
\cline{2-10}
 & \cite{Sharma2021} & 2021 & Detection & Remote Patient Monitoring (RPM) & GAF & SVM\par{}KNN & SVM and KNN with different Fusion Methods & Accuracy\par{}Precision\par{}Recall\par{}Sensitivity\par{}Specificity & Accuracy: \par{}96.33\% \\* 
\cline{2-10}

& \cite{ChenandDeng2021} & 2021 & Prediction & Indoor Speed Estimation & STFT & CNN-LSTM & Other Papers & Accuracy & Accuracy:\par{}96.33\% \\* 
\cline{2-10}

 & \cite{Singh2021} & 2021 & Classification & Brain Signal Classification & STFT & VGG-16 & SVM\par{}Logistic Regression\par{}Random Forest & Accuracy\par{}Specificity\par{}Sensitivity\par{}Precision & Accuracy:\par{}88.04\% \\* 
\cline{2-10}
 & \cite{ZhangandHang2022} & 2022 & Classification & ECG Signal Classification & RP\par{}FT & CNN-LSTM & Other Papers & Accuracy\par{}Specificity\par{}Sensitivity & Accuracy: 99.06\%\par{}Sensitivity : 98.29\%\par{}Specificity: 99.73\% \\* 
\cline{2-10}
 & \cite{Dou2022} & 2022 & Classification & ECG Signal Classification & CWT & GoogLeNet & AlexNet\par{}VGGNet & Accuracy\par{}Specificity\par{}Sensitivity & Accuracy:\par{}94.28\% \\* 
\cline{2-10}
 & \cite{Abdel-Basset2022}& 2022 & Classification & HAR & RP\par{}MTF\par{}GAF & CNN-based model & Other Papers & Accuracy\par{}F1-score\par{}Precision\par{}Recall & Accuracy:\par{}HHAR --\textgreater{} 98.90\%\par{}MEHEALTH --\textgreater{}99.68\% \\* 
\cline{2-10}
 &  \cite{DePaula2023} & 2023 & Classification & EEG Signal Classification & GADF\par{}GASF\par{}MTF\par{}RP & ImageNet\par{}DenseNet\par{}ResNet\par{}Google Net\par{}AlexNet & 1D-kernel-based CNNs & Accuracy & Accuracy: (ResNet50) \par{}
RP --\textgreater{} 96\%\par{}
GADF --\textgreater{} 94\%\par{}
MTF --\textgreater{} 88\%\par{}
GASF --\textgreater{} 54\%
  \\* 
\hline
\multirow{8}{0.071\linewidth}{\hspace{0pt}{}Industrial} &  \cite{Ferraro2020} & 2020 & Prediction & Maintenance Prediction & GAF & CNN & LSTM & Accuracy\par{}Precision\par{}Recall & Accuracy:\par{}97.70\% \\* 
\cline{2-10}
 & \cite{JiangandYen2021}& 2021 & Prediction & Product Quality\par{}Prediction & MTF & CNN-LSTM & Other Papers\par{}DNN\par{}MC-DNN & MAPE & MAPE:\par{}3-state MTF --\textgreater{}3.11\%\par{}4-state MTF --\textgreater{}2.94\%\par{}5-state MTF --\textgreater{}3.24\% \\* 
\cline{2-10}
 & \cite{Garcia2020}& 2021 & Detection & Anomaly Detection & GAF\par{}MTF\par{}RP\par{}GS\par{}STFT\par{}DWT & CNN & - & F1-score\par{}AUC\par{}True and False Positive Rates (TPR, FPR) & AUC:\par{}SC -\textgreater{} 92\par{}GS -\textgreater{} 89\par{}MTF Mod. -\textgreater{} 87\par{}GAF Mod. -\textgreater{} 84 \\* 
\cline{2-10}
 & \cite{DeSanto2022} & 2022 & Prediction & Maintenance Prediction & RP\par{}GAF\par{}MTF \par{}CWT & CNN & LSTM\par{}GRU\par{}XGBoost\par{}ResNet-50\par{}DenseNet-121\par{} VGG-16 & F1-score\par{}Precision\par{}Recall & F1-score: \par{}GAN--\textgreater{} 34.47\par{}CNN--\textgreater{} 59.24\par{}Accuracy:\textbf{}RP --\textgreater{} 0.95 \\* 
\cline{2-10}
 &  \cite{Wang2022} & 2022 & Detection & Bearig Fault Detection & STFT\par{}The direct drawing method\par{} GADF \par{}MTF & VGG & - & Accuracy\par{}Loss Function\par{}Confusion Matrix & Accuracy:\par{}
MFPT :\par{}
STFT --\textgreater{}99.8\%\par{}
CWRU:\par{}
DDM --\textgreater{} 93.8\%\par{}
GADF --\textgreater{} 78.1\%\par{}
MTF --\textgreater{} 79.7\%\par{}
STFT --\textgreater{}  100\%
 \\* 
\cline{2-10}
 & \cite{Bai2022}& 2022 & Detection & Fault Diagnosis & GAF\par{}MTF & DCGAN\par{}EfficientNet & CNN\par{}VGG16\par{}GoogleNet & Accuracy & Accuracy: \par{}0.8541 \\* 
\cline{2-10}
 & \cite{Sun2023}& 2023 & Detection & Fault Diagnosis & GAF\par{}MTF & ResNet & DCNN & Accuracy & Accuracy: \par{}72.14\% \\* 
\cline{2-10}
 & \cite{ZhangandShang2023} & 2023 & Detection & Fault Diagnosis & GASF\par{}GADF & ELM \par{}CNN & AlexNet\par{}Neural Network (NN)\par{}SVM\par{}KNN & Accuracy\par{}Precision\par{}Recall\par{}F1-measure & Accuracy: \par{}99.2\%  \\
\hline

\end{tabular}
  \end{threeparttable}
\end{table*}

\begin{table*}
\centering
\begin{threeparttable}
\caption* {\textbf{Table \ref{table:SummaryStudies} Continued:} Summary of Image Transformation Techniques Studies in IoT Application Domain} 
\begin{tabular}{>{\hspace{0pt}}m{0.080\linewidth}>
{\centering\hspace{0pt}}m{0.031\linewidth}>
{\centering\hspace{0pt}}m{0.032\linewidth}>
{\centering\hspace{0pt}}m{0.093\linewidth}>
{\centering\hspace{0pt}}m{0.090\linewidth}>
{\centering\hspace{0pt}}m{0.059\linewidth}>
{\centering\hspace{0pt}}m{0.10\linewidth}>
{\centering\hspace{0pt}}m{0.08\linewidth}>
{\centering\hspace{0pt}}m{0.06\linewidth}>
{\centering\arraybackslash\hspace{0pt}}m{0.14\linewidth}} 

\\\hline
\textbf{Domain} & \textbf{Ref.} & \textbf{Year} & \textbf{Problem Type} & \textbf{Application Type} & \textbf{Methods} & \textbf{Models} & \textbf{Comparision Models} & \multicolumn{1}{>{\centering\hspace{0pt}}m{0.06\linewidth}}{\textbf{Performance}\par{}\textbf{ Metrics}} & \textbf{Results} \\* 
\hline
\begin{tabular}{@{}l@{}}Environmental \\ Monitoring\end{tabular} & \cite{Abidi2023}& 2023 & Classification & Time-series\par{}Classification & \par{}GADF\par{}GASF\par{}MTF\par{}RP & CNN\par{}ResNet-50 & Other Papers & Accuracy\par{}F1-Socore\par{}Kappa & F1-scores: \par{}Reunion Island --\textgreater{} 89.34\%, \par{}Dordogne --\textgreater{} 90.26\% \par{}Koumbia study --\textgreater{} 78.94\% \\ 
\hline

\begin{tabular}{@{}l@{}} Smart \\ Building \end{tabular} & \cite{Sayed2023} & 2023 & Prediction & Building Occupancy Prediction & Data Normalization \par{} Matrix Conversion & CNN & KNN\par{}DT\par{}RF & Accuracy\par{}Precision\par{}Recall\par{}F1 Score\par{}Kappa & Accuracy:\par{}
SRD --\textgreater{} 99.11\%\par{}
LRD --\textgreater{}  98.54\%\par{}
UCI --\textgreater{} 99.42\%  \\ 
\hline
\begin{tabular}{@{}l@{}}Transportation \\ and Logistics\end{tabular}& \cite{Huang2023} & 2021 & Imputation & Traffic Data Imputation & GASF & DCGAN & Other Papers & MAE\par{}RMSE\par{}MRE & MAE: \par{}13.7\% \\ 
\hline

\begin{tabular}{@{}l@{}}Wearable  \\ Devices\end{tabular}&  \cite{Xu2020b} & 2020 & Classification & HAR & GASF\par{}GADF & Multi-dilated Kernel Residual (Mdk-Res) Module\par{}ResNet & Multilayer Perceptron (MLP)\par{}LSTM\par{}CNN\_1D\par{}CNN\_2D\par{}ResNet\par{}GoogleNet & Accuracy\par{}Precision\par{}Recall\par{}F-measure & Accuracy:\par{}Proposed --\textgreater{} 96.83\%\par{}CNN--\textgreater{} 93.23\par{}LSTM --\textgreater{}87.53 \\* 
\hline
\multirow{4}{0.071\linewidth}{\hspace{0pt}{}\begin{tabular}[c]{@{}c@{}}Others  \footnotemark[1]\end{tabular}} & \cite{Yang2019} & 2019 & Classification & Time-series\par{}Classification & GASF\par{}GADF\par{}MTF & ConvNet & ConvNet\par{}VGG16\par{}Other Papers & Error Rate & Error rates:\par{}Wafer --\textgreater{} 0.4 for MTF\par{}ECG --\textgreater{} 5.35 for GADF \\* 
\cline{2-10}
 & \cite{JiangandZhang2021} & 2021 & Classification & Time-series\par{}Classification & GAF\par{}MTF & ADDN & ResNet Encoder\par{}MLP\par{}MCDCNN\par{}Time-CNN & Accuracy\par{}Arithmetic Rank\par{}Geometric Rank\par{}Mean Per Class Error (MPCE) & MPCE:\par{}
UTS --\textgreater{} 2.90\par{}
MTS --\textgreater{} 4.00
  \\* 
\cline{2-10}
 &  \cite{Quan2023}& 2023 & Classification & Time-series\par{}Classification & GAF\par{}MTF\par{}RP\par{}GMR & ResNet &  ResNet\par{}
Dynamic Time Warping (DTW)\par{}
MLP\par{}
Fully Convolutional Network (FCN)
 & Error Rate & Error rates: \par{}
GMR --\textgreater{} 0.2305 \par{}
GAF --\textgreater{} 0.2431\par{}
MTF --\textgreater{} 0.2863\par{}
RP --\textgreater{} 0.2543
 \\* 
\cline{2-10}
 & \cite{Hasan2023} & 2023 & Detection & Sensor Fault Diagnosis & GAF & ResNet18- SVM-GAN & ResNet18-SVM & Accuracy\par{}F1-score\par{}Precision\par{}Recall\par{}Confusion Matrix & Accuracy: \par{}98.7\% \\
\hline

\end{tabular}
\begin{tablenotes}
      \item[*] {The studies have not been provided with any domain-specific information.}
    \end{tablenotes}
  \end{threeparttable}
\end{table*}

 \normalsize
\subsection{Transportation and Logistics}

IoT applications in transportation and logistics include fleet management, vehicle tracking, route optimization, cargo monitoring, and driver safety. These applications have the potential to transform the industry by enabling intelligent decision-making, reducing costs, and improving customer experiences.

Huang et al. \cite{Huang2023} developed a new method, namely The Traffic Sensor Data Imputation GAN (TSDIGAN), for missing data reconstruction.  GASF was employed in the paper to process time-series traffic data and transform it into an image format for missing value imputation using CNN.

\subsection{Wearable Devices}

Wearable devices focus on the integration of technology into portable devices that individuals can wear. These devices are equipped with sensors, connectivity capabilities, and computing power, enabling them to collect data, interact with the environment, and provide personalized experiences.

Wearable devices incorporate various sensors to collect data about the user and their environment, such as accelerometers, heart rate monitors, GPS, temperature sensors, etc. They are also connected to other devices or networks through wireless technologies such as Bluetooth and Wi-Fi. Thus, wearable devices offer individuals convenient access to personalized data and experiences, empowering them to monitor their health, improve their fitness, and stay connected in a more seamless and unobtrusive manner.

With the advancement of the IoT and wearable devices, sensor-based HAR has gained importance due to convenience and privacy characteristics. Xu et al. \cite{Xu2020b} presented two improvements based on GAF and deep CNN for HAR. The findings indicated that the developed model was able to efficiently extract multi-scale features and improve the accuracy of activity recognition by utilizing the GAF algorithm's characteristics, along with the structure and advantages of CNN, residual learning, and dilated convolution.

\subsection{Others}
Beyond the above-mentioned IoT domains, various studies employ data from different fields within IoT. In these studies, the effects of the proposed methods on the datasets obtained from diverse fields were examined. For example, Yang et al. \cite{Yang2019} used two well-known MTS datasets, Wafer and ECG, to classify 1D signals. MTS data was transformed into 2D images by applying GASF, GADF, and MTF. These images were then concatenated as RGB input channels for the ConvNet classification model. The study concluded that the choice of encoding methods had no significant impact on the prediction results. 
Jiang et al. \cite{JiangandZhang2021} evaluated the Adaptive Dila-DenseNet (ADDN) model for classifying UTS and MTS data across 24 benchmark IoT datasets. Both UTS and MTS data were converted into GM-images by leveraging GAF and MTF methods to feed into the ADDN model. 
Quan et al. \cite{Quan2023} investigated the impact of different feature construction and fusion methods on time-series classification results. They proposed an improved Multi-Scale ResNet (MSResNet) for time-series classification. In this study, three images encoded with different methods, including GAF, MTF, and RP, were superimposed as three-channel data inputs as GMR images. Besides, 1D multi-scale features and 2D image features were fused using two distinct methods, including Squeeze-and-Excitation (SE) and Self-Attention (SA) feature fusion architectures.
Hasan et al. \cite{Hasan2023} introduced a sensor fault detection approach based on digital twins. They used the GAN method to create the digital representation of the sensor. Also, The GAN was trained with images obtained by converting time-series using GAF.

\normalsize

\section{RESEARCH CHALLENGES AND FUTURE  DIRECTIONS}

Converting time-series data into images attracted significant attention in facilitating IoT data analysis. However, this transformation process has a set of challenges. The major challenges and potential solutions are presented to address them as follows for researchers \cite{Fu2024, GuoandWan2019, Ghahramani2022, Ghasemieh2022, Yu2022, Debayle2018, Yuan2021, Kiangala2020, Ashraf2023, JiangandLiu2022, Estebsari2020, Nizam2022, Ali2023}.

\begin{itemize}

\item IoT time-series data is often prone to noise and missing values caused by sensor failures or network problems, which can adversely affect the image quality. Also, missing data when creating an image can lead to incomplete representations. Imputing methods such as image inpainting models \cite{Fu2024} or GAN-based models \cite{GuoandWan2019} can be utilized to handle missing gaps.

\item Encoding large-scale IoT time-series data can be computationally expensive and memory-intensive. In order to capture meaningful patterns without overwhelming the computational resources, optimal image dimensions should be determined \cite{Ghasemieh2022}. On the other hand, it should be noted that very small sizes can lead to the loss of essential details while reducing memory and computational costs. Moreover, current techniques can be redesigned to accelerate image conversion.

\item The process of transforming time-series data into images involves compressing the temporal information into a 2D representation. This compression can cause information loss. Balancing the trade-off between dimensional reduction and information loss is a critical challenge in this field \cite{Yu2022}. To minimize information loss, researchers should focus on developing robust transformation techniques that balance dimensionality reduction and information preservation. Besides, in order to avoid information loss, modifications can be made to the transformation methods, such as changing the function in a formula \cite{Debayle2018}, \cite{Yuan2021}.

\item IoT time-series data can involve multiple variables or sensors, resulting in MTS. However, the methods described cannot be applied directly to MTS. To deal with this issue, various approaches can be developed, such as adapting methods to MTS data or transforming MTS data to the appropriate format for the methods. Dimension reduction methods can be utilized to implement encoding techniques directly \cite{Kiangala2020,Ashraf2023}. Also, effective fusion techniques can be developed to transform MTS data \cite{JiangandLiu2022}. 

\item  Real-time or near-real-time image representations for dynamic IoT environments can be challenging \cite{Estebsari2020}. If the image transformation process takes longer than the time between intervals, the system can fail eventually. This delay may be unacceptable for decision-making systems within a short period in IoT applications such as IIoT and smart cities. Increasing hardware resources can help reduce computation time. Also, edge devices provide prominent computational resources for faster real-time decision-making \cite{Nizam2022}. Furthermore, combining edge and cloud architecture in IoT can effectively address network traffic congestion and latency concerns \cite{Ali2023}.

\end{itemize}

\section{CONCLUSION}

In recent years, the transformation of time-series data into images has become widespread. However, the adoption of these techniques in IoT domains is still in its early stages, with expectations for them to become commonplace across most IoT domains in the near future.
This study presents a comprehensive review of image transformation techniques employed in various IoT domains, including smart buildings, industrial settings, energy management, healthcare, security, and more. We categorize existing studies based on their encoding techniques, IoT application areas, and data types. 
In the literature, various transformation techniques are applied to both univariate and multivariate time-series IoT data. These transformation techniques are typically used in conjunction with fusion techniques for multivariate time-series IoT data. 
Among the techniques employed, GAF and MTF are the most commonly used image transformation techniques, particularly in domains such as energy management, healthcare, and industrial applications with purposes such as anomaly detection, fault diagnosis, and time-series classification. 
Additionally, this paper discusses the associated challenges, open issues, and future research directions.


\end{document}